\documentclass[runningheads,a4paper]{llncs}

\usepackage{amsmath,amssymb} 
\usepackage{color}           
\usepackage{graphicx}
\usepackage{caption}
\usepackage{epstopdf}        
\usepackage{url}
\usepackage[ruled,linesnumbered]{algorithm2e}
\usepackage{enumerate}
\urldef{\mailsa}\path|{vbehzadan, amunir}@unr.edu|
\newcommand{\keywords}[1]{\par\addvspace\baselineskip
\noindent\keywordname\enspace\ignorespaces#1}

\begin{document}

\mainmatter  

\title{Vulnerability of Deep Reinforcement Learning to Policy Induction Attacks}

\titlerunning{Vulnerability of Deep Reinforcement Learning to Policy Induction Attacks}

%
%
\author{Vahid Behzadan
\and Arslan Munir}
\authorrunning{Vulnerability of Deep Reinforcement Learning to Policy Induction Attacks}

\institute{Department of Computer Science and Engineering\\ University of Nevada, Reno\\
1664 N Virginia St, Reno, NV 89557\\
\mailsa\\
}
%
%

\toctitle{Lecture Notes in Computer Science}
\tocauthor{Authors' Instructions}
\maketitle

\begin{abstract}
Deep learning classifiers are known to be inherently vulnerable to manipulation by intentionally perturbed inputs, named adversarial examples. In this work, we establish that reinforcement learning techniques based on Deep Q-Networks (DQNs) are also vulnerable to adversarial input perturbations, and verify the transferability of adversarial examples across different DQN models. Furthermore, we present a novel class of attacks based on this vulnerability that enable policy manipulation and induction in the learning process of DQNs. We propose an attack mechanism that exploits the transferability of adversarial examples to implement policy induction attacks on DQNs, and demonstrate its efficacy and impact through experimental study of a game-learning scenario. 
\end{abstract}

\section{Introduction}\label{intro}
Inspired by the psychological and neuroscientific models of natural learning, Reinforcement Learning (RL) techniques aim to optimize the actions of intelligent agents in complex environments by learning effective controls and reactions that maximize the long-term reward of agents. \cite{sutton1998introduction}. The applications of RL range from combinatorial search problems such as learning to play games \cite{ghory2004reinforcement} to autonomous navigation \cite{dai2005approach}, multi-agent systems \cite{busoniu2008comprehensive}, and optimal control \cite{sutton1992reinforcement}. However, classic RL techniques generally rely on hand-crafted representations of sensory input, thus limiting their performance in the complex and high-dimensional real world environments. To overcome this limitation, recent developments combine RL techniques with the significant feature extraction and processing capabilities of deep learning models in a framework known as Deep Q-Network (DQN) \cite{mnih2013playing}. This approach exploits deep neural networks for both feature selection and Q-function approximation, hence enabling unprecedented performance in complex settings such as learning efficient playing strategies from unlabeled video frames of Atari games \cite{mnih2015human}, robotic manipulation \cite{gu2016deep}, and autonomous navigation of aerial \cite{zhang2015learning} and ground vehicles \cite{hussein2016deep}.

The growing interest in the application of DQNs in critical systems necessitate the investigation of this framework with regards to its resilience and robustness to adversarial attacks on the integrity of reinforcement learning processes. The reliance of RL on interactions with the environment gives rise to an inherent vulnerability which makes the process of learning susceptible to perturbation as a result of changes in the observable environment. Exploiting this vulnerability provides adversaries with the means to disrupt or change control policies, leading to unintended and potentially harmful actions. For instance, manipulation of the obstacle avoidance and navigation policies learned by autonomous Unmanned Aerial Vehicles (UAV) enables the adversary to use such systems as kinetic weapons by inducing actions that lead to intentional collisions.

In this paper, we study the efficacy and impact of policy induction attacks on the Deep Q-Learning RL framework. To this end, we propose a novel attack methodology based on adversarial example attacks against deep learning models \cite{papernot2016limitations}. Through experimental results, we verify that similar to classifiers, Q networks are also vulnerable to adversarial examples, and confirm the transferability of such examples between different models. We then evaluate the proposed attack methodology on the original DQN architecture of Mnih, et. al. \cite{mnih2015human}, the results of which verify the feasibility of policy induction attacks by incurring minimal perturbations in the environment or sensory inputs of an RL system. We also discuss the insufficiency of defensive distillation \cite{papernot2015distillation} and adversarial training \cite{carlini2016towards} techniques as state of the art countermeasures proposed against adversarial example attacks on deep learning classifiers, and present potential techniques to mitigate the effect of policy induction attacks against DQNs.

The remainder of this paper is organized as follows: Section \ref{related} presents an overview of Q-Learning, Deep Q-Networks, and adversarial examples. Section \ref{model} formalizes the problem and defines the target and attacker models. In Section \ref{method}, we outline the attack methodology and algorithm, followed by the experimental evaluation of the proposed methodology in Section \ref{results}. A high-level discussion on effectiveness of the current countermeasures is presented in Section \ref{counter}, and the paper is concluded in Section \ref{conclusion} with remarks on future research directions.

\section{Background}\label{related}

\subsection{Q-Learning}

The generic RL problem can be formally modeled as a Markov Decision Process, described by the tuple $MDP = (S, A, P, R)$, where $S$ is the set of reachable states in the process, $A$ is the set of available actions, $R$ is the mapping of transitions to the immediate reward, and $P$ represents the transition probabilities. At any given time-step $t$, the MDP is at a state $s_t\in S$. The RL agent's choice of action at time $t$, $a_t \in A$ causes a transition from $s_t$ to a state $s_{t+1}$ according to the transition probability $P_{s_t , s_{t+a}}^{a_t}$. The agent receives a reward $r_t = R(s_t, a_t) \in \mathbb{R}$ for choosing the action $a_t$ at state $s_t$.

Interactions of the agent with MDP are captured in a policy $\pi$. When such interactions are deterministic, the policy $\pi: S\rightarrow A$ is a mapping between the states and their corresponding actions. A stochastic policy $\pi(s,a)$ represents the probability of optimality for action $a$ at state $s$.

The objective of RL is to find the optimal policy $\pi^\ast$ that maximizes the cumulative reward over time at time $t$, denoted by the return function $\hat{R} = \sum_{T}^{t' = t} \gamma^{t'-t} r_{t'}$, where $\gamma < 1$ is the discount factor representing the diminishing worth of rewards obtained further in time, hence ensuring that $\hat{R}$ is bounded.

One approach to this problem is to estimate the optimal value of each action, defined as the expected sum of future rewards when taking that action and following the optimal policy thereafter. The value of an action $a$ in a state $s$ is given by the action-value function $Q$ defined as:
\begin{eqnarray} \label{bellman}
Q(s,a) = R(s, a) + \gamma max_{a'}(Q(s',a'))
\end{eqnarray}

Where $s'$ is the state that emerges as a result of action $a$, and $a'$ is a possible action in state $s'$. The optimal $Q$ value given a policy $pi$ is hence defined as: $Q^\ast (s, a) = max_{\pi} Q^{\pi} (s, a)$, and the optimal policy is given by $\pi^\ast(s) = \arg\max_a Q(s,a)$

The Q-learning method estimates the optimal action policies by using the Bellman equation $Q_{i+1} (s,a) = \mathbf{E}[R + \gamma \max_a Q_i]$ as the iterative update of a value iteration technique. Practical implementation of Q-learning is commonly based on function approximation of the parametrized Q-function $Q(s,a; \theta) \approx Q^\ast (s,a)$. A common technique for approximating the parametrized non-linear Q-function is to train a neural network whose weights correspond to $\theta$. Such neural networks, commonly referred to as Q-networks, are trained such that at every iteration $i$, it minimizes the loss function 
\begin{eqnarray}
L_i(\theta_i) = \mathbf{E}_{s, a\sim \rho(.)} [(y_i - Q(s,a,;\theta_i))^2]
\end{eqnarray}

where $y_i = \mathbf{E}[R + \gamma \max_{a'}Q(s',a';\theta_{i-1}) | s,a]$, and $\rho(s,a)$ is a probability distribution over states $s$ and actions $a$. This optimization problem is typically solved using computationally efficient techniques such as Stochastic Gradient Descent (SGD) \cite{baird1999gradient}.
%
%
%
%
%
%
%

\subsection{Deep Q Networks}

\begin{figure}
	
	\centering
	
	\includegraphics[scale=0.40]{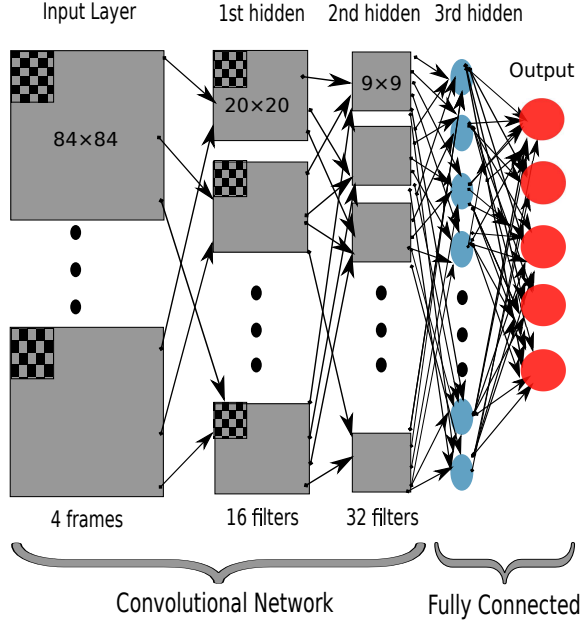}
	
	\caption{DQN architecture for end-to-end learning of Atari 2600 game plays}
	
	\label{fig:deepq}
	
\end{figure}

Classical Q-networks present a number of major disadvantages in the Q-learning process. First, the sequential processing of consecutive observations breaks the \emph{iid} requirement of training data as successive samples are correlated. Furthermore, slight changes to Q-values leads to rapid changes in the policy estimated by Q-network, thus enabling policy oscillations. Also, since the scale of rewards and Q-values are unknown, the gradients of Q-networks can be sufficiently large to render the backpropagation process unstable.

A deep Q network (DQN) \cite{mnih2013playing} is a multi-layered Q-network designed to mitigate such disadvantages. To overcome the issue of correlation between consecutive observations, DQN employs a technique named \emph{experience replay}: Instead of training on successive observations, experience replay samples a random batch of previous observations stored in the replay memory to train on. As a result, the correlation between successive training samples is broken and the iid setting is re-established. In order to avoid oscillations, DQN fixes the parameters of the optimization target $y_i$. These parameters are then updated at regulat intervals by adopting the current weights of the Q-network. The issue of unstability in backpropagation is also solved in DQN by clipping the reward values to the range $[-1,+1]$, thus preventing Q-values from becoming too large.

Mnih et. al. \cite{mnih2015human} demonstrate the application of this new Q-network technique to end-to-end learning of Q values in playing Atari games based on observations of pixel values in the game environtment. The neural network architecture of this work is depicted in figure \ref{fig:deepq}. To capture the movements in the game environment, Mnih et. al. use stacks of 4 consecutive image frames as the input to the network. To train the network, a random batch is sampled from the previous observation tuples $(s_t, a_t, r_t, s_{t+1})$. Each observation is then processed by 2 layers of convolutional neural networks to learn the features of input images, which are then employed by feed-forward layers to approximate the Q-function. The target network $\hat{Q}$, with parameters $\theta^{-}$, is synchronized with the parameters of the original $Q$ network at fixed periods intervals. i.e., at every $i$th iteration,  $\theta^-_{t} = \theta_t$, and is kept fixed until the next synchronization. The target value for optimization of DQN learning thus becomes:

\begin{eqnarray}
y'_t \equiv r_{t+1} + \gamma max_{a'} \hat{Q}(S_{t+1}, a'; \theta^-)
\end{eqnarray}

Accordingly, the training process can be stated as:

\begin{eqnarray}\label{SGD}
min_{a_t} (y'_t - Q(s_t, a_t, \theta))^2
\end{eqnarray}

\subsection{Adversarial Examples}

in \cite{szegedy2013intriguing}, Szegedy et. al. report an intriguing discovery: several machine learning models, including deep neural networks, are vulnerable to adversarial examples. That is, these machine learning models misclassify inputs that are only slightly different from correctly classified samples drawn from the data distribution. Furthermore, a wide variety of models with different architectures trained on different subsets of the training data misclassify the same adversarial example.

This suggests that adversarial examples expose fundamental blind spots in machine learning algorithms. The issue can be stated as follows: Consider a machine learning system $M$ and a benign input sample $C$ which is correctly classified by the machine learning system, i.e. $M(C) = y_{true}$. According to the report of Szegedy \cite{szegedy2013intriguing} and many proceeding studies \cite{papernot2016limitations}, it is possible to construct an adversarial example $A = C + \delta$, which is perceptually indistinguishable from $C$, but is classified incorrectly, i.e. $M(A) \neq y_{true}$.

Adversarial examples are misclassified far more often than examples that have been perturbed by random noise, even if the magnitude of the noise is much larger than the magnitude of the adversarial perturbation \cite{goodfellow2014explaining}. According to the objective of adversaries, adversarial example attacks are generally classified into the following two categories: 

\begin{enumerate}
	
	\item Misclassification attacks, which aim for generating examples that are classified incorrectly by the target network
	
	\item Targeted attacks, whose goal is to generate samples that the target misclassifies into an arbitrary class designated by the attacker.
	
\end{enumerate}

To generate such adversarial examples, several algorithms have been proposed, such as the Fast Gradient Sign Method (FGSM) by Goodfellow et. al., \cite{goodfellow2014explaining}, and the Jacobian Saliency Map Algorithm (JSMA) approach by Papernot et. al., \cite{papernot2016limitations}. A grounding assumption in many of the crafting algorithms is that the attacker has complete knowledge of the target neural networks such as its architecture, weights, and other hyperparameters. Recently, Papernot et. al. \cite{papernot2016practical} proposed the first black-box approach to generating adversarial examples. This method exploits the generalized nature of adversarial examples: an adversarial example generated for a neural network classifier applies to most other neural network classifiers that perform the same classification task, regardless of their architecture, parameters, and even the distribution of training data. Accordingly, the approach of \cite{papernot2016practical} is based on generating a replica of the target network. To train this replica, the attacker creates and trains over a dataset from a mixture of samples obtained by observing target's performance, and synthetically generated inputs and label pairs. Once trained, any of the adversarial example crafting algorithms that require knowledge of the target network can be applied to the replica. Due to the transferability of adversarial examples, the perturbed samples generated from the replica network will induce misclassifications in many of the other networks that perform the same task. In the following sections, we describe how a similar approach can be adopted in policy induction attacks against DQNs.

\section{Threat Model}\label{model}
We consider an attacker whose goal is to perturb the optimality of actions taken by a DQN learner via inducing an arbitrary policy $\pi_{adv}$ on the target DQN. The attacker is assumed to have minimal \emph{a priori} information of the target, such as the type and format of inputs to the DQN, as well as its reward function $R$ and an estimate for the frequency of updating the $\hat{Q}$ network. It is noteworthy that even if the target's reward function is not known, it can be estimated via Inverse Reinforcement Learning techniques \cite{gao2012survey}. No knowledge of the target's exact architecture is considered in this work, but the attacker can estimate this architecture based on the conventions applied to the input type (e.g. image and video input may indicate a convolutional neural network, speech and voice data point towards a recurrent neural network, etc.). 

In this model, the attacker has no direct influence on the target's architecture and parameters, including its reward function and the optimization mechanism. The only parameter that the attacker can directly manipulate is the configuration of the environment observed by the target. For instance, in the case of video game learning \cite{mnih2013playing}, the attacker is capable of changing the pixel values of the game's frames, but not the score. In cyber-physical scenarios, such perturbations can be implemented by strategic rearrangement of objects or precise illumination of certain areas via tools such as laser pointers. To this end, we assume that the attacker is capable of changing the state before it is observed by the target, either by predicting future states, or after such states are generated by the environment's dynamics. The latter can be achieved if the attacker has a faster action speed than the target's sampling rate, or by inducing a delay between generation of the new environment and its observation by the target. 

To avoid detection and minimize influence on the environment's dynamics, we impose an extra constraint on the attack such that the magnitude of perturbations applied in each configuration must be smaller than a set value denoted by $\epsilon$. Also, we do not limit the attacker's domain of perturbations (e.g. in the case of video games, the attacker may change the value of any pixel at any position on the screen).

\section{Attack Mechanism}\label{method}
As discussed in Section \ref{related}, the DQN framework of Mnih et. al. \cite{mnih2015human} can be seen as consisting of two neural networks, one is the native network which performs the image classification and function approximation, and the other is the auxiliary $\hat{Q}$ network whose architecture and parameters are copies of the native network sampled once every $c$ iterations. Training of DQN is performed optimizing the loss function of equation \ref{SGD} by Stochastic Gradient Descent (SGD). Due to the similarity of this process and the training mechanism of neural network classifiers, we hypothesize that the function approximators of DQN are also vulnerable to adversarial example attacks. In other words, the set of all possible inputs to the approximated functions $Q$ and $\hat{Q}$ contains elements which cause the approximated functions to generate outputs that are different from the output of the original $Q$ function. Furthermore, we hypothesize that similar to the case of classifiers, the elements that cause one DQN to generate incorrect $Q$ values will incur the same effect on other DQNs that approximate the same Q-function. 

Consequently, the attacker can manipulate a DQN's learning process by crafting states $s_t$ such that $\hat{Q}(s_{t+1}, a; \theta^-_{t})$ identifies an incorrect choice of optimal action at $s_{t+1}$. If the attacker is capable of crafting adversarial inputs $s'_t$ and $s'_{t+1}$ such that the value of Equation \ref{SGD} is minimized for a specific action $a'$, then the policy learned by DQN at this time-step is optimized towards suggesting $a'$ as the optimal action given the state $s_t$.

Considering that the attacker is not aware of the target's network architecture and its parameters at every time step, crafting adversarial states must rely on black-box techniques such as those introduced in \cite{papernot2016practical}. Attacker can exploit the transferability of adversarial examples by obtaining the state perturbations from a replica of the target's DQN. At every time step of training this replica, attacker calculates the perturbation vectors $\hat{\delta}_{t+1}$ for the next state $s_{t+1}$ such that $max_{a'} \hat{Q}(s_{t+1} + \hat{\delta}_{t+1}, a'; \theta^-_{t})$ causes $\hat{Q}$ to generate its maximum when $a' = {\pi^\ast_{adv}}(s_{t+1})$, i.e., the maximum reward at the next state is obtained when the optimal action taken at that state is determined by attacker's policy. 

This is procedurally similar to targeted misclassification attacks described in Section \ref{related} that aim to find minimal perturbations to an input sample such that the classifier assigns the maximum value of likelihood to an incorrect target class. Therefore, the adversarial example crafting techniques developed for classifiers, such as the Fast Gradient Sign Method (FGSM) and the Jacobian Saliency Map Algorithm (JSMA), can be applied to obtain the perturbation vector $\hat{\delta}_{t+1}$. 

\begin{figure}
	
	\centering
	
	\includegraphics[scale=0.29]{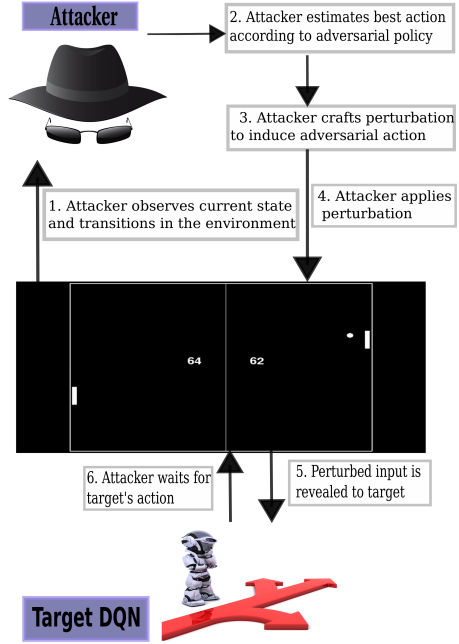}
	
	\caption{Exploitation cycle of policy induction attack}
	
	\label{fig:exploit}
	
\end{figure}

The procedure of this attack can be divided into the two phases of initialization and exploitation. The initialization phase implements processes that must be performed before the target begins interacting with the environment, which are:

\begin{enumerate}
	\item Train a DQN based on attacker's reward function $r'$ to obtain the adversarial policy $\pi^\ast_{adv}$
	\item Create a replica of the target's DQN and initialize with random parameters 
\end{enumerate}

The exploitation phase implements the attack processes such as crafting adversarial inputs. This phase constitutes an attack cycle depicted in figure \ref{fig:exploit}. The cycle initiates with the attacker's first observation of the environment, and runs in tandem with the target's operation. Algorithm \ref{algo:exploit} details the procedural flow of this phase.

\begin{algorithm}[t]
	\fontsize{7.5pt}{7.5pt}
	\caption{Exploitation Procedure}
	\label{algo:exploit}
	\SetKwInOut{Input}{input}\SetKwInOut{Output}{output}
	\Input{adversarial policy $\pi^*_{adv}$, initialized replica DQNs $Q'$, $\hat{Q'}$, synchronization frequency $c$, number of iterations N}
	
	\For{observation = 1, N}{
		Observe current state $s_t$, action $a_t$, reward $r_t$, and resulting state $s_{t+1}$ \\
		\If {$s_{t+1}$ is not terminal}
		{
			set $a'_{adv} = \pi^*_{adv}(s_{t+1})$ \\
			Calculate perturbation vector $\hat{\delta}_{t+1} = Craft(\hat{Q'}, a'_{adv}, s_{t+1} )$\\
			Update $s_{t+1} \leftarrow s_{t+1} + \hat{\delta}_{t+1}$ \\
			Set $y_t = (r_t + max_{a'} \hat{Q'}(s_{t+1} + \hat{\delta}_{t+1}, a'; \theta'_{-})$\\
			Perform SGD on $(y_t - Q'(s_t, a_t, \theta'))^2$ w.r.t $\theta'$\\
		}
		Reveal $s_{t+1}$ to target \\
		\lIf {$observation\mod c = 0$}
		{
			$\theta'_{-} \leftarrow \theta'$
		}
	}
\end{algorithm}

\section{Experimental Verification}\label{results}
\begin{figure}[!h]
	
	\centering
	
	\includegraphics[scale=0.36]{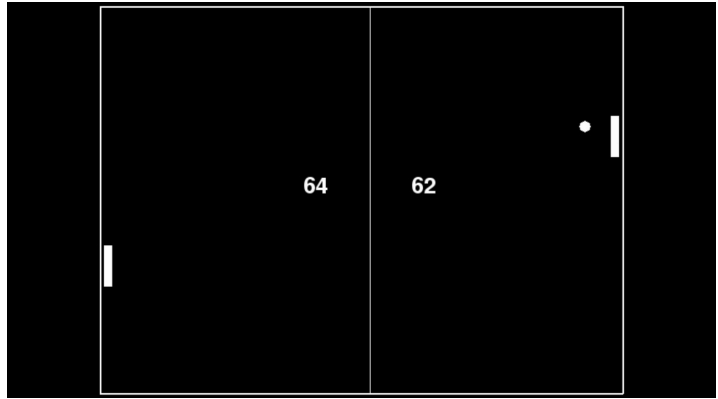}
	
	\caption{Game of Pong}
	
	\label{fig:pong}
	
\end{figure}

To study the performance and efficacy of the proposed mechanism, we examine the targeting of Mnih et. al.'s DQN designed to learn Atari 2600 games \cite{mnih2015human}. In our setup, we train the network on a game of Pong implemented in Python using the PyGame library \cite{mcgugan2007beginning}. The game is played against an opponent with a modest level of heuristic artificial intelligence, and is customized to handle the delays in DQN's reaction due to the training process. The game's backened provides the DQN agent with the game screen sampled at 8Hz, as well as the game score (+1 for win, -1 for lose, 0 for ongoing game) throughout each episode of the game. The set of available actions $A = \{UP, DOWN, Stand\}$ enables the DQN agent to control the movements of its paddle. Figure \ref{fig:pong} illustrates the game screen of Pong used in our experiments.

The training process of DQN is implemented in TensorFlow \cite{abadi2016tensorflow} and executed on an Amazon EC2 g2.2xlarge instance \cite{gilani2015application} with 8 Intel Xeon E5-2670 CPU cores and a NVIDIA GPU with 1536 CUDA cores and 4GB of video memory. Each state observed by the DQN is a stack of 4 consecutive 80x80 gray-scale game frames. Similar to the original architecture of Mnih et. al. \cite{mnih2015human}, this input is first passed through two convolutional layers to extract a compressed feature space for the following two feed-forward layers for Q function estimation. The discount factor $\gamma$ is set to $0.99$, and the initial probability of taking a random action is set to $1$, which is annealed after every $500000$ actions. The agent is also set to train its DQN after every $50000$ observations. Regular training of this DQN takes approximately 1.5 million iterations ($\sim$16 hours on the g2.2xlarge instance) to reach a winning average of 51\% against the heuristic AI of its opponent\footnote{As expected, longer training of this DQN leads to better results. After a 2-week period of training we verified the convergent trait of our implementation by witnessing winning averages of more than 80\%.}

Following the threat model presented in Section \ref{model}, this experiment considers an attacker capable of observing the states interactions between his target DQN and the game, but his domain of influence is limited to implementation of minor changes on the environment. Considering the visual representation of the environment in this setup, the minor changes incurred by attacker take the form of perturbing pixel values in the 4 consecutive frames of a given state. 

\subsection{Evaluation of Vulnerability to Adversarial Examples}
\begin{figure}[h]
	
	\centering
	
	\includegraphics[scale=0.60]{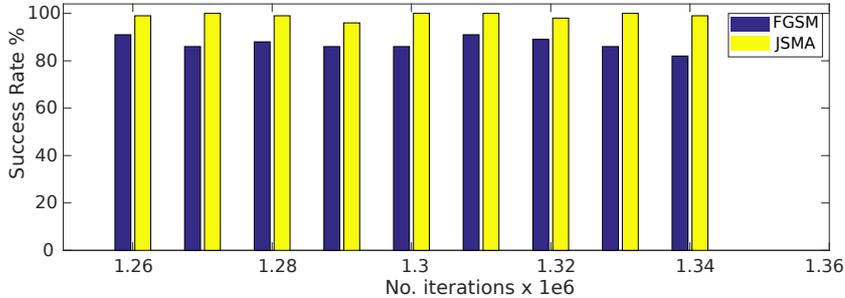}
	
	\caption{Success rate of crafting adversarial examples for DQN}
	
	\label{fig:plt1}
	
\end{figure}

Successful implementations of the proposed policy induction attack mechanisms rely on the vulnerability of DQNs to targeted adversarial perturbations. To verify the existence of this vulnerability, the $\hat{Q}$ networks of target were sampled at regular intervals during training in the game environment. In the next step, 100 observations comprised of a pair of consecutive states $(s_{t},s_{t+1})$  were randomly selected from the experience memory of DQN, to ensure the possibility of their occurrence in the game. Considering $s_{t+1}$ to be the variable that can be manipulated by the attacker, it is passed along with the model $\hat{Q }$ to the adversarial example crafting algorithms. To study the extent of vulnerability, we evaluated the success rate of both FGSM and JSMA algorithms for each of the 100 random observations in inducing a random game action other than the current optimal $a^\ast_t$ . The results, presented in Figure \ref{fig:plt1}, verify that DQNs are indeed vulnerable to adversarial example attacks. It is noteworthy that the success rate of FGSM with a fixed perturbation limit decreases by one percent per 100000 observations as the number of observations increases. Yet, JSMA seems to be more robust to this effect as it maintains a success rate of ~100 percent throughout the experiment.

\subsection{Verification of Transferability}
\begin{figure}[h]
	
	\centering
	
	\includegraphics[width=\linewidth]{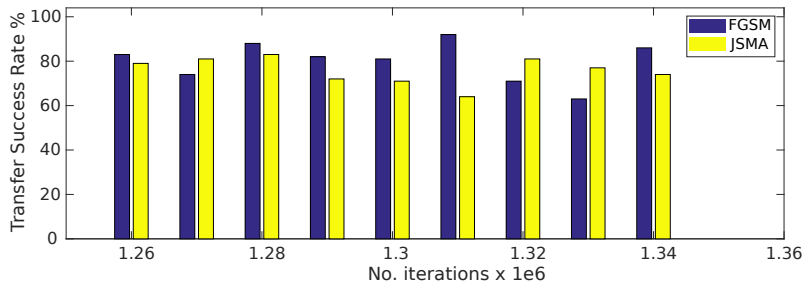}
	
	\caption{Transferability of adversarial examples in DQN}
	
	\label{fig:plt2}
	
\end{figure}

To measure the transferability of adversarial examples between models, we trained another Q-network with a similar architecture on the same experience memory of the game at the sampled instances of the previous experiment. It is noteworthy that due to random initializations, the exploration mechanism, and the stochastic nature of SGD, even similar Q-networks trained on the same set of observations will obtain different sets of weights. The second Q-network was tested to measure its vulnerability to the adversarial examples obtained from the last experiment. Figure \ref{fig:plt2} shows that more than $70\%$ of the perturbations obtained from both FGSM and JSMA methods also affect the second network, hence verifying the transferability of adversarial examples between DQNs.

\subsection{Performance of Proposed Policy Induction Attack}
Our final experiment tests the performance of our proposed exploitation mechanism. In this experiment, we consider an adversary whose reward value is the exact opposite of the game score, meaning that it aims to devise a policy that maximizes the number of lost games. To obtain this policy, we trained an adversarial DQN on the game, whose reward value was the negative of the value obtained from target DQN's reward function. With the adversarial policy at hand, a target DQN was setup to train on the game environment to maximize the original reward function. The game environment was modified to allow perturbation of pixel values in game frames by the adversary. A second DQN was also setup to train on the target's observations to provide an estimation of the target DQN to enable blackbox crafting of adversarial example. At every observation, the adversarial policy obtained in the initialization phase was consulted to calculate the action that would satisfy the adversary's goal. Then, the JSMA algorithm was utilized to generate the adversarial example that would cause the output of the replica DQN network to be the action selected by the adversarial policy. This example was then passed to the target DQN as its observation.
\begin{figure}[h]
	
	\centering
	
	\includegraphics[width=\linewidth]{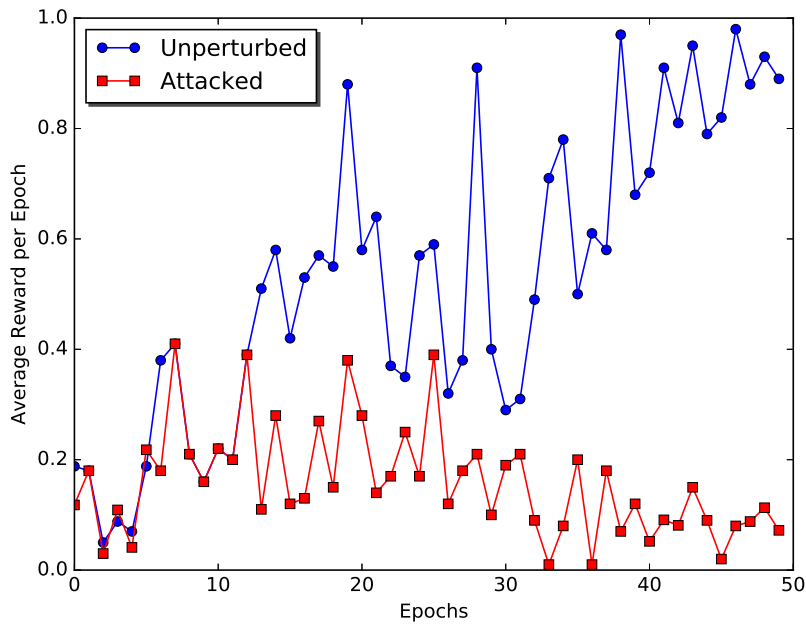}
	
	\caption{Comparison of rewards between unperturbed and attacked DQNs}
	
	\label{fig:plt3}
	
\end{figure}
Figure \ref{fig:plt3} compares the performance of unperturbed and attacked DQNs in terms of their reward values, measured as the difference of current game score with the average score. It can be seen that the reward value for the targeted agent rapidly falls below the unperturbed case and maintains the trend of losing the game throughout the experiment. This result confirms the efficacy of our proposed attack mechanism, and verifies the vulnerability of Deep Q-Networks to policy induction attacks.

\section{Discussion on Current Counter-Measures}\label{counter}
Since the introduction of adversarial examples by Szgedey, et. al. \cite{szegedy2013intriguing}, various counter-measures have been proposed to mitigate the exploitation of this vulnerability in deep neural networks. Goodfellow et. al. \cite{goodfellow2014explaining} proposed to retrain deep networks on a set of minimally perturbed adversarial examples to prevent their misclassification. This approach suffers from two inherent short-comings: Firstly, it aims to increase the amount of perturbations required to craft an adversarial example. Second, this approach does not provide a comprehensive counter-measure as it is computationally inefficient to find all possible adversarial examples. Furthermore, Papernot et. al. \cite{papernot2016practical} argue that by training the network on adversarial examples, the emerging network will have new adversarial examples and hence this technique does not solve the problem of exploiting this vulnerability for critical systems. Consequently, Papernot, et. al \cite{papernot2015distillation} proposed a technique named Defensive Distillation, which is also based on retraining the network on a dimensionally-reduced set of training data. This approach, too, was recently shown to be insufficient in mitigating adversarial examples \cite{carlini2016defensive}. It is hence concluded that the current state of the art in countering adversarial examples and their exploitation is incapable of providing a concrete defense against such exploitations.

In the context of policy induction attacks, we conjecture that the temporal features of the training process may be utilized to provide protection mechanisms. The proposed attack mechanism relies on the assumption that due to the decreasing chance of random actions, the target DQN is most likely to perform the action induced by adversarial inputs as the number of iterations progress. This may be mitigated by implementing adaptive exploration-exploitation mechanisms that both increase and decrease the chance of random actions according to the performance of the trained model. Also, it may be possible to exploit spatio-temporal pattern recognition techniques to detect and omit regular perturbations during the pre-processing phase of the learning process. Investigating such techniques is the priority of our future work.

\section{Conclusions and Future Work}\label{conclusion}
We established the vulnerability of reinforcement learning based on Deep Q-Networks to policy induction attacks. Furthermore, we proposed an attack mechanism which exploits the vulnerability of deep neural networks to adversarial examples, and demonstrated its efficacy and impact through experiments on a game-learning DQN.

This preliminary work solicitates a wide-range of studies on the security of Deep Reinforcement Learning. As discussed in Section \ref{counter}, novel countermeasures need to be investigated to mitigate the effect of such attacks on DQNs deployed in cyber-physical and critical systems. Also, an analytical treatment of the problem to establish the bounds and relationships of model parameters, such as network architecture and exploration mechanisms, with DQN's vulnerability to policy induction will provide deeper insight and guidelines into designing safe and secure deep reinforcement learning architectures.

\end{document}